\documentclass[letterpaper]{article} 
\usepackage{aaai24}
\usepackage{times}  
\usepackage{helvet}  
\usepackage{courier}  
\usepackage[hyphens]{url}  
\usepackage{graphicx} 
\usepackage{natbib}  
\usepackage{caption} 
\usepackage{algorithm}
\usepackage{algorithmic}

\usepackage{booktabs}
\usepackage{multirow}
\usepackage{colortbl}
\usepackage{amsmath}
\usepackage{amssymb}

\usepackage{newfloat}
\usepackage{listings}
\DeclareCaptionStyle{ruled}{labelfont=normalfont,labelsep=colon,strut=off} 
\floatstyle{ruled}
\newfloat{listing}{tb}{lst}{}
\floatname{listing}{Listing}
\title{ROIC-DM: Robust Text Inference and Classification via Diffusion Model}

\author {
    Shilong Yuan\textsuperscript{\rm 1},
    Wei Yuan\textsuperscript{\rm 2},
    Hongzhi Yin\textsuperscript{\rm 2}
    Tieke He\textsuperscript{\rm 1}\thanks{Corresponding author.}
}
\affiliations {
    \textsuperscript{\rm 1}Nanjing University\\
    \textsuperscript{\rm 2}The University of Queensland\\
    shilongyuan@nju.edu.cn, w.yuan@uq.edu.au,  db.hongzhi@gmail.com, hetieke@nju.edu.cn
}

\usepackage{bibentry}

\begin{document}

\maketitle
\begin{abstract}

While language models have made many milestones in text inference and classification tasks, they remain susceptible to adversarial attacks that can lead to unforeseen outcomes. Existing works alleviate this problem by equipping language models with defense patches. However, these defense strategies often rely on impractical assumptions or entail substantial sacrifices in model performance. Consequently, enhancing the resilience of the target model using such defense mechanisms is a formidable challenge.
This paper introduces an innovative model for robust text inference and classification, built upon diffusion models (ROIC-DM). Benefiting from its training involving denoising stages, ROIC-DM inherently exhibits greater robustness compared to conventional language models. Moreover, ROIC-DM can attain comparable, and in some cases, superior performance to language models, by effectively incorporating them as advisory components.
Extensive experiments conducted with several strong textual adversarial attacks on three datasets demonstrate that (1) ROIC-DM outperforms traditional language models in robustness, even when the latter are fortified with advanced defense mechanisms; (2) ROIC-DM can achieve comparable and even better performance than traditional language models by using them as advisors.

\end{abstract}

\section{Introduction}
Text inference and classification are two fundamental and significant tasks in Natural Language Processing (NLP)~\cite{li2022survey}.
In recent years, large language models have made impressive advancements in these two tasks, however, these models are pointed out to be extremely vulnerable to textual adversarial attacks~\cite{papernot2016limitations},
wherein adversaries can easily compromise the performance of these models by crafting deceptive inputs.
To address this issue, many defense methods have been proposed to improve the adversarial robustness of these language models. These defense methods can generally be classified into three categories: adversarial detection~\cite{c:5,c:6}, adversarial training~\cite{wang2021infobert, Zhu2020FreeLB, madry2018towards}, and adversarial purification~\cite{c:4}.

Although these defense approaches can slightly alleviate the threats of adversarial attacks to some extent, they either rely on strong assumptions or sacrifice too much model performance, limiting their practical usage~\cite{c:4}.
Specifically, adversarial detection~\cite{c:5,mosca2022suspicious} and adversarial training methods~\cite{Zhu2020FreeLB,madry2018towards,wang2021adversarial} require prior knowledge of adversarial samples.
However, in real-world scenarios, these adversarial samples are unavailable before potential attacks are launched. Even when attacks have been executed, differentiating and collecting the adversarial samples from input data remains challenging since the perturbations in these samples are imperceptible.
Adversarial purification methods can avoid this dilemma as they can purify adversarial text without requiring knowledge of the specific attacks. Nevertheless, since purification operations have to be applied to all input data, these methods will deteriorate the model's performance, as the original information of clean data will be inevitably modified during the purification process. Achieving the removal of adversarial perturbations while preserving the original meanings of the data remains a challenging task for these purification methods. 

Considering the limitations of existing defense methods, it is not easy to improve the adversarial robustness of existing language models by equipping them with certain defending patches. In light of this, this paper proposes to explore a new kind of text inference and classification model that is robust enough against most adversarial attacks. 

Diffusion models, as a new kind of generative models, have exhibited powerful learning ability in the computer vision community~\cite{kong2021diffwave,ramesh2022hierarchical,dhariwal2021diffusion}.
Naturally, some researchers attempt to transplant them into natural language processing areas~\cite{DBLP:journals/corr/abs-2205-14217,gong2023diffuseq,savinov2022stepunrolled, NEURIPS2021_958c5305}. But all these works are focusing on text generation tasks since diffusion models are generally used as generative models.

\begin{figure*}[t]
\centering
\includegraphics[width=1\textwidth]{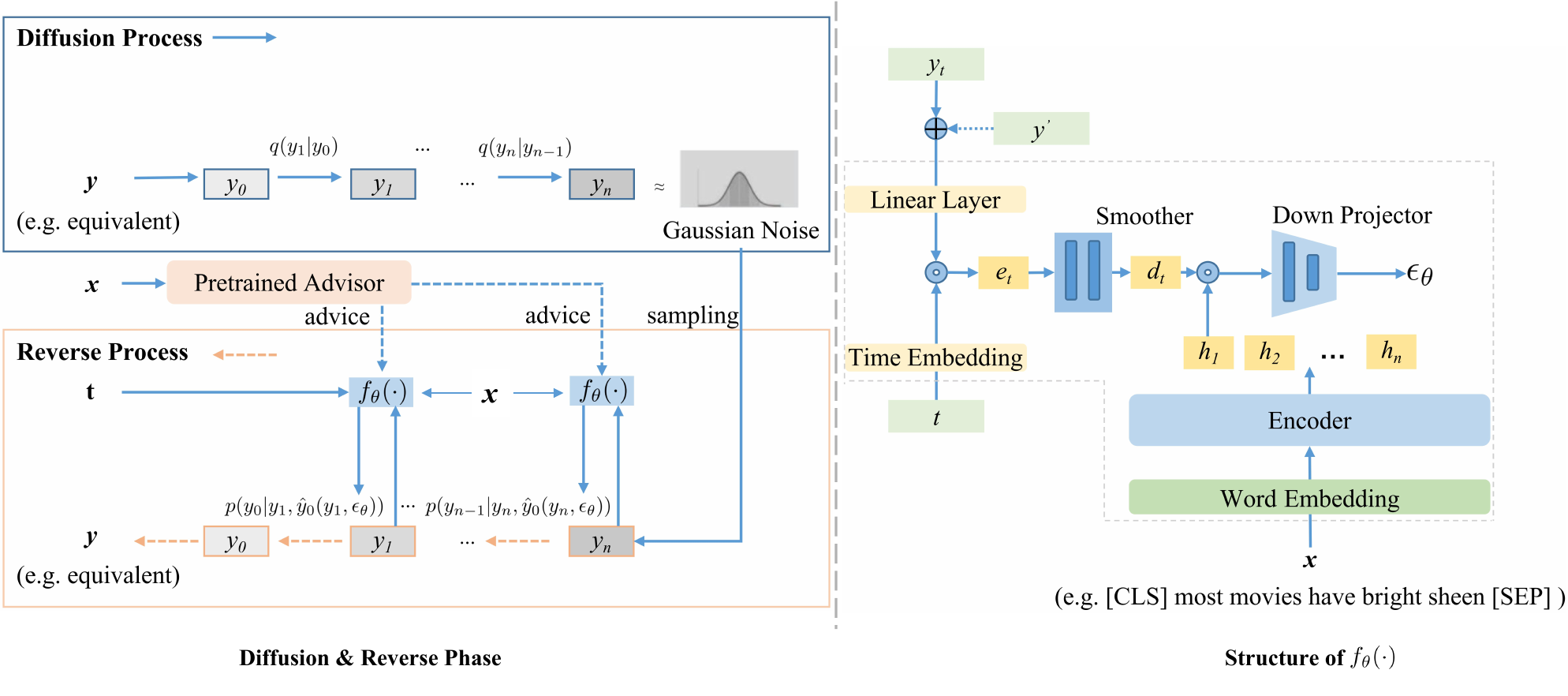} 
\caption{ Illustration of our proposed Robust text Inference and Classification Diffusion Model (ROIC-DM). The figure on the left illustrates the diffusion phase and the reverse phase. The figure on the right illustrates the architecture of noise estimator $f_{\theta}$. $\mathbf{x}$ is the input text; $\mathbf{y}$ is the categorical label; $t$ is the time step number; $\epsilon_{\theta}$ is the predicted noise. The $\odot$ is the element-wise product. The $\oplus$ indicates element-wise summation.}
\label{fig1}
\end{figure*}

In this paper, we take the first step to investigate the usage of diffusion models in text inference and classification contexts. The basic motivation for replacing the traditional language models with diffusion models is that the diffusion models contain the diffusion-denoising steps so that they themselves would be more robust as they can more accurately estimate the whole data space~\cite{chen2023robust}.
The left part of Figure~\ref{fig1} explains our motivation. During the reverse process, the model $f_{\theta}$ will ``match" the text $\mathbf{x}$ with many polluted $\mathbf{y}_{t}$ until it removes all the noises. As a result, the data space of $(x,y)$ pairs is large and $f_{\theta}$ would be robust to adversarial perturbations.
However, applying diffusion models in text inference and classification tasks faces many challenges. 
Firstly, diffusion models are originally generative models, how to employ them for text classification and inference is non-trivial.
Besides, the classical language models have been developed for a long duration in text classification and inference tasks with tons of researchers' efforts, therefore, naively leveraging the new style of models in these fields may fail to achieve comparable performance. It is significant to incorporate the previous language models' abilities in the diffusion models to gain better performance.

To address the above challenges, we propose ROIC-DM (\emph{Ro}bost text \emph{i}nference and \emph{c}lassification \emph{d}iffusion \emph{m}odel), which is the first text inference and classification diffusion model.
ROIC-DM first modifies the original diffusion models to make them suitable for text classification and inference tasks. 
Then, to further improve the effectiveness and efficiency, ROIC-DM incorporates the traditional language models as advisors to provide advice during the denoising process. 
It is worth noting that ROIC-DM is much different from text purification~\cite{c:4}, as the latter only uses diffusion models for data preprocessing while ROIC-DM directly utilizes diffusion models to solve the tasks.
Extensive experiments on three real-world datasets demonstrate that (1) ROIC-DM is more robust to adversarial attacks than existing language models even when they are equipped with advanced defense methods. (2) ROIC-DM can achieve comparable and even better performance than traditional language models by using them as advisors.
The main contributions of this paper are as follows.
\begin{itemize}
    \item To the best of our knowledge, we are the first to introduce diffusion models in text classification and inference fields.
    \item We propose ROIC-DM which achieves better performance and robustness than language models.
    \item We conduct extensive experiments on three real-world datasets including AG NEWS, SST, and MRPC. The experimental results showcase the effectiveness and robustness of our proposed methods.
\end{itemize}
\section{Related Work}
\subsection{Textual Adversarial Attacks}
Although language models have made great achievements, they are revealed to be vulnerable to adversarial attacks, i.e., these language models can be easily manipulated by adversaries via minor revisions of normal samples.
Generally, most existing textual adversarial attacks achieve their malicious goals by replacing specific words in input texts~\cite{alzantot2018generating,jin2020bert,ren2019generating}.
These methods usually assume that the model is black-box but the logits of the output prediction are available. Then, they attempt to design some strategies to find appropriate words to replace.
For example, ~\cite{jin2020bert,mrkvsic2016counter,ren2019generating} replace words with synonyms, while \cite{jin2020bert,li2020bert} uses greedy-search methods to get substitutes.

\subsection{Textual Adversarial Defenses}
To improve the robustness of these language models, many defenses are proposed.
These defense approaches can be generally classified into adversarial training, adversarial detection, and adversarial purification.
The line of adversarial training~\cite{wang2021infobert, wang2020cat,Zhu2020FreeLB,zhou2019learning,madry2018towards} is to incorporate perturbations during a model's training process so that the model can be robust to the potential risks. The works of adversarial detection~\cite{c:5,c:6} aim to filter out the adversarial samples.
The purification methods~\cite{samangouei2018defense,li-etal-2023-text} employ generative models to purify adversarial inputs before feeding them to a model. However, all these defense methods have certain limitations. To be specific, the adversarial training methods and detection approaches require prior knowledge of attacks which is infeasible in practice, while the adversarial purification methods cannot avoid the modification of normal inputs. As a result, using defense methods to improve language models' robustness is challenging.

\section{Preliminaries: Diffusion Models}\label{rw_dm}
In this section, we introduce the theory of diffusion model~\cite{sohl2015deep,ho2020denoising,song2020denoising}. Generally, a diffusion model processes data in two steps. First, it gradually transforms raw data into a Gaussian distribution through a diffusion process. Then, It learns the reverse procedure to reconstruct the original data from Gaussian white noise.~\cite{sohl2015deep}. 
The following paragraphs formally describe these two processes.

In \textbf{diffusion process},  the diffusion model incrementally corrupts the original representation $\mathbf{x}_{0}$ into a Gaussian noise $\mathbf{x}_{t}$ via a Markov Chain with fixed parameters~\cite{ho2020denoising} in \textit{T} steps:
\begin{equation}
    q(\mathbf{x}_t|\mathbf{x}_{t-1})=\mathcal{N}\left(\mathbf{x}_t;\sqrt{1-\beta_t}\mathbf{x}_{t-1}, \beta_t\mathbf{I}\right)
    \label{eq:diffu_step}
\end{equation}
where $\mathcal{N}\left(x;\mu,\sigma^{2}\right)$ represents $x$ sampled from a Gaussian distribution with a mean $\mu$ and variance $\sigma^{2}$.
The value of $\beta_{t} $ is determined by a pre-defined noise schedule $\beta$ that regulates the amount of noise injected at each step.
The common noise schedules encompass square-root~\cite{li2022diffusion}, cosine~\cite{ho2020denoising}, and linear~\cite{nichol2021improved} functions. According to~\cite{c:11}, $\mathbf{x}_{t}$ can be directly computed conditioned on $\mathbf{x}_{0}$ with the following transformation:
\begin{align}
    q(\mathbf{x}_t|\mathbf{x}_0) &= \mathcal{N} \left(\mathbf{x}_t;\sqrt{\overline{\alpha}_t}\mathbf{x}_0,(1-\overline{\alpha}_t)\mathbf{I}\right)\\
    \overline{\alpha}_t&=\prod_{i=1}^t\alpha_i,~~\alpha_i=1-\beta_i.
\end{align}
With the trick of re-parameter, we can get $\mathbf{x}_{t}$ by adding Guassian noise $\epsilon\sim\mathcal{N}(0,\mathbf{I})$ to $\mathbf{x}_{0}$ as follows:
\begin{equation}
\mathbf{x}_t=\sqrt{\bar{\alpha}_t}\mathbf{x}_0+\sqrt{1-\bar{\alpha}_t}\epsilon\label{eq:diffu}
\end{equation}

The \textbf{reverse process} is a Markov Chain with a learnable $\theta$ to denoise $\mathbf{x}_{T}$ to $\mathbf{x}_{0}$. Specifically, when provided with the current representation $\mathbf{x}_{s}$, the subsequent representation $\mathbf{x}_{s-1}$ after denoising is calculated as follows: 
\begin{align}
    p(\mathbf{x}_{s-1}|\mathbf{x}_s,\mathbf{x}_0)&=\mathcal{N}\left(\mathbf{x}_{s-1};\Tilde{\mathbf{\mu}}_s(\mathbf{x}_s,\mathbf{x}_0),\Tilde{\beta}_s\mathbf{I}\right)\label{eq:reversion}\\
\Tilde{\mathbf{\mu}}_s(\mathbf{x}_s,\mathbf{x}_0)&=\frac{\sqrt{\overline{\alpha}_{s-1}}\beta_s}{1-\overline{\alpha}_s}\mathbf{x}_0+\frac{\sqrt{\alpha_s}(1-\overline{\alpha}_{s-1})}{1-\overline{\alpha}_s}\mathbf{x}_s \\ \Tilde{\beta_s}&=\frac{1-\overline{\alpha}_{s-1}}{1-\overline{\alpha}_s}\beta_s 
\end{align}

Since $\mathbf{x}_{0}$ is unknown during the reverse phase, the diffusion model utilizes a noise estimator $f_{\theta}(\mathbf{x}_{t},t)$, which is typically modelled by a deep neural network, such as Transformer~\cite{vaswani2017attention} or U-Net~\cite{ronneberger2015u}. 
We optimize the reverse phase, by optimizing the variational lower bound(VLB)
\begin{align}
\mathcal{L}_{vlb}&= \underbrace{\mathbb{E}_q\left[D_{KL}\left(q(\mathbf{x}_t|\mathbf{x}_0)||p(\mathbf{x}_t)\right)\right]}_{L_t}\nonumber\\
&+\underbrace{\mathbb{E}_q\left[\sum_{s=2}^t D_{KL}(q(\mathbf{x}_{s-1}|\mathbf{x}_s,\mathbf{x}_0)||p_{\theta}(\mathbf{x}_{s-1}|\mathbf{x}_s))\right]}_{L_{t-1}}\nonumber\\
&-\underbrace{\textrm{log} p_{\theta}(\mathbf{x}_0|\mathbf{x}_1)}_{L_0}
\label{eq:loss_diffu}
\end{align}

Finally, after simplifying~\cite{kingma2021variational}, the diffusion loss is as follows:

\begin{equation}
    \mathcal{L}_{simple} = \mathbb{E}_{t,\mathbf{x}_0,\mathbf{\epsilon}}\left[||\mathbf{\epsilon}-f_{\theta}(\mathbf{x}_t,t)||^2\right]
    \label{eq:obj_simple}
\end{equation}
More details of the general diffusion model can be found in previous works~\cite{ho2020denoising,song2020denoising}.

\section{Methodology}
In this section, we first introduce how to apply diffusion models in text classification and inference, i.e., ROIC-DM.
Then, we describe how to incorporate the knowledge from pre-trained models to ROIC-DM to improve the model performance.
Finally, we present the detailed model architecture of ROIC-DM.
The overview of our ROIC-DM is illustrated in Figure~\ref{fig1}.

\subsection{Diffusion Model for Text Classification and Inference}
The skeleton of ROIC-DM is the diffusion model.
As diffusion models exhibit strong learning ability in Computer Vision, many works attempt to employ them to solve NLP tasks~\cite{li2022diffusion,gong2022diffuseq}. 
However, all of these works mainly utilize diffusion models to solve natural language generation problems since diffusion models are a kind of generative model.
This paper takes the first step to modifying diffusion models as robust classifiers to solve text classification and inference problems.
The technical details are as follows.

In the diffusion process, ROIC-DM gradually adds noise to the label $\mathbf{y}_{0}$. Note that since the label $\mathbf{y}_{0}$ usually can be denoted as a one-hot vector, ROIC-DM can directly add noise to it~\cite{hoogeboom2021argmax,c:10}. Therefore, the only difference in the diffusion process between ROIC-DM and other standard diffusion models is that our input is a label while others are images or sentences.
As a result, the diffused $\mathbf{y}$ in each timestep can be obtained as follows:
\begin{equation}\label{eq_get_yt}
\mathbf{y}_t=\sqrt{\bar{\alpha}_t}\mathbf{y}_0+\sqrt{1-\bar{\alpha}_t}\epsilon \quad \epsilon\sim\mathcal{N}(\mathbf{0},\mathbf{I})
\end{equation}

\begin{algorithm}[!htbp]
  \renewcommand{\algorithmicrequire}{\textbf{Input:}}
  \renewcommand{\algorithmicensure}{\textbf{Output:}}
	\caption{Training procedure of ROIC-DM}
	\label{alg:diffusion}
	\begin{algorithmic}[1]
            \REQUIRE text $x$, learning epochs $E$, maximum diffusion steps $T$, linear schedule $\beta$, model $f_\theta(\cdot)$, advisor prediction $\mathbf{y}^{\prime}$ (optional), $\dots$
            \ENSURE well trained $f_\theta(\cdot)$;
		\WHILE{$j < E$}
            \STATE $t\sim\mathrm{uniform}(\{0,\ldots,T\})$;
            \STATE $\mathbf{y}_t,\epsilon\leftarrow$ E.q.~\ref{eq_get_yt}
            \IF{use advisor}
            \STATE $\epsilon_\theta= f_\theta(\mathbf{x},\mathbf{y}_t,t,\mathbf{y}^{\prime}) $
            \ELSE
            \STATE $\epsilon_\theta= f_\theta(\mathbf{x},\mathbf{y}_t,t)$
            \ENDIF
            \STATE update $f_{\theta}(\cdot)$ using $\nabla_{\theta}\left\|\epsilon-\epsilon_{\theta}\right\|^{2}$\;
            \STATE $j = j + 1$;
            \ENDWHILE
	\end{algorithmic} 
    \label{alg-1}
\end{algorithm}

The critical difference between ROIC-DM and the generative diffusion models is the reverse process.
Specifically, for traditional diffusion models, during their reverse process, they utilize a noise estimator $f_{\theta}(\mathbf{x}_{t},t)$ to predict the noise (note that the $\mathbf{x}_{t}$ in this case is equal to $\mathbf{y}_{t}$ in ROIC-DM). 
However, in ROIC-DM, directly denoising $\mathbf{y}_{t}$ without any conditions is meaningless since it is just a certain category label~\cite{hoogeboom2021argmax,c:10}.
Thus, ROIC-DM constructs a trainable model $f_{\theta}(\mathbf{x},\mathbf{y}_t,t)$ whose goal is to generate a noise $\epsilon_{\theta}$ to recover $\mathbf{y}_t$ to $\mathbf{y}_{t-1}$ considering the corresponding text context $\mathbf{x}$.

Algorithm~\ref{alg-1} shows how to train the model $f_{\theta}(\mathbf{x},\mathbf{y}_t,t)$.
To be specific, ROIC-DM randomly selects a pair of data $(\mathbf{x},\mathbf{y})$ and calculates $\mathbf{y}_{t}$ using E.q.~\ref{eq_get_yt}. Then, it predicts the noise using $f_{\theta}(\mathbf{x},\mathbf{y}_t,t)$ and calculates the loss with E.q.~\ref{eq_41_loss}. Above steps iteratively continue until model convergence.
\begin{equation}
    \mathcal{L} = \mathbb{E}_{t,\mathbf{x}_0,\mathbf{\epsilon}}\left[||\mathbf{\epsilon}-f_{\theta}(\mathbf{x},\mathbf{y}_t,t)||^2\right]
    \label{eq_41_loss}
\end{equation}

The inference procedure of ROIC-DM is displayed in Algorithm~\ref{alg-2}. In the inference phase, ROIC-DM randomly samples a Gaussian noise $\mathbf{y}_{T}$ as the start point. Then, ROIC-DM recovers $\mathbf{y}_{T}$ to $\mathbf{y}_{0}$ by repeating the following equations:
\begin{equation}\label{eq_gety0}
    \begin{aligned}
        \mathbf{y}_{t-1}&=\tilde{\mu}(\hat{\mathbf{y}_{0}},\mathbf{y}_{t})+\tilde{\beta}_{t}*\zeta \\
\tilde{\mu}_t\left(\mathbf{y}_t,\mathbf{\hat{y}}_0\right)&=\frac{\sqrt{\overline{\alpha}_{t-1}}\beta_t}{1-\overline{\alpha}_t}\mathbf{\hat{y}_0}+\frac{\sqrt{\alpha_t}(1-\overline{\alpha}_{t-1})}{1-\overline{\alpha}_t}\mathbf{y}_t \\
    \tilde{\beta_t}&=\frac{1-\overline{\alpha}_{t-1}}{1-\overline{\alpha}_t}\beta_t\\
\hat{\mathbf{y}}_0&=\frac1{\sqrt{\bar{\alpha}_t}}\left(\mathbf{y}_t-\sqrt{1-\bar{\alpha}_t}\mathbf{\epsilon}_\theta\right)
    \end{aligned}
\end{equation}
where $\zeta\sim\mathcal{N}(\mathbf{0},\mathbf{I})$, $\epsilon_{\theta}$ is calculated by well-trained $f_{\theta}(\cdot)$.
After obtaining $\mathbf{y}_{0}$, ROIC-DM selects the position with the maximum value as the final predicted class for $\mathbf{x}$.

\begin{algorithm}[H]
	\renewcommand{\algorithmicrequire}{\textbf{Input:}}
  \renewcommand{\algorithmicensure}{\textbf{Output:}}
	\caption{Inference procedure of ROIC-DM}
	\label{alg:revision}
	\begin{algorithmic}[1]
            \REQUIRE text $x$, total reverse steps $T$, linear schedule $\beta$, model $f_\theta(\cdot)$, advisor prediction $\mathbf{y}^{\prime} $ $\dots$
            \ENSURE $label$ 
            \STATE sample $\mathbf{y}_T\sim \mathcal{N}(\mathbf{0}, \mathbf{I})$
            \STATE $t=T$
		\WHILE{$t > 0$}
            \STATE sample $\zeta\sim\mathcal{N}(\mathbf{0},\mathbf{I})$
            \IF{use advisor}
            \STATE $\epsilon_\theta = f_\theta(\mathbf{x},\mathbf{y}_t,t,\mathbf{y}^{\prime}) $
            \ELSE
            \STATE $\epsilon_\theta = f_\theta(\mathbf{x},\mathbf{y}_t,t)$
            \ENDIF
            \STATE $ \mathbf{y}_{t-1}\leftarrow$ calculate E.q.~\ref{eq_gety0} with $\zeta$
            \STATE $t = t -1$
            \ENDWHILE
            \STATE $label\leftarrow$find the position with max value in $\mathbf{y}_{0}$
	\end{algorithmic}  
     \label{alg-2}
\end{algorithm}

\subsection{Pre-trained Advisor Improved ROIC-DM}
Although language models have been revealed to be vulnerable to adversarial attacks, they have developed for a long time and have achieved remarkable results in text classification and inference tasks, especially the large pre-trained language models~\cite{devlin2018bert,minaee2021deep}. Therefore, it is necessary to build our ROIC-DM based on the power of pre-trained language models.

To achieve that, before training ROIC-DM, we first fine-tune a large pre-trained language model (e.g., BERT~\cite{devlin2018bert}) on the target classification dataset.
Then, during the reverse process, ROIC-DM generates $\epsilon_{\theta}$ not only considering $\mathbf{x}$, but also the prediction from the fine-tuned pre-trained model. 
Specifically, we transfer the knowledge of fine-tuned model to ROIC-DM by utilizing its soft-label $\mathbf{y}^{\prime}$ based on the input $\mathbf{x}$, which is inspired by the observations from knowledge distillation~\cite{gou2021knowledge} said that model's soft label contains many useful auxiliary information.
In ROIC-DM, we incorporate the fine-tuned model's knowledge by directly adding ${\mathbf{y}^{\prime}}$ to $\mathbf{y}_{t}$.
The left part of Figure~\ref{fig1} shows the framework of our ROIC-DM.

\subsection{Model Architecture of ${f_\theta(\cdot)}$}\label{ms}
The right part of Figure~\ref{fig1} presents the model architecture of the noise estimator $f_{\theta}(\cdot)$ in ROIC-DM.
Generally, $f_{\theta}$ contains an encoder to extract features from the context $\mathbf{x}$, a time embedding table to encode the timestep, a normalization layer to make the output be more smooth, and a down projector to output the noise.

\subsubsection{Encoder}
We use the encoder block of the BERT~\cite{devlin2018bert} as the feature extractor to convert the words to hidden states. 
Then, we leverage the hidden state of $[CLS]$ token\footnote{We also tried using the average of hidden states as the feature vector and obtain similar results.}, which is appended at the start of the text, as the feature vector of text $\mathbf{x}$, i.e., $\mathbf{h}_{1}$ in Figure~\ref{fig1}. 
\begin{equation}
    \mathbf{h}_{1}\leftarrow Encoder(\mathbf{x})
\end{equation}

\subsubsection{Time Embedding and Linear Layer}
The diffusion step $t$ is uniformly sampled randomly from $\{0,\ldots, T\}$.
Then, we leverage a time embedding table to learn the features of time steps $\mathbf{t}$~\cite{xiao2021tackling}.
For $\mathbf{y}_{t}$, we utilize a full-connected layer to transform it to the same size as $\mathbf{t}$ and then conduct element-wise product to fuse the information of $\mathbf{t}$ and $\mathbf{y}_{t}$.
\begin{equation}
    \mathbf{e}_{t} = \mathbf{t}\odot Linear(\mathbf{y}_{t})
\end{equation}

\subsubsection{Smoother} 
Smoother is consisted of a softmax function and a layer normalization. Such a combination can be used in certain layers of neural networks to enhance the network's representational capacity and convergence performance and lead to improved performance and faster training speed in learning tasks~\cite{huang2023normalization}. Here, we use this module to process the fused vector $\mathbf{e}_{t}$.
\begin{equation}
    \mathbf{d}_{t}= LN(Softplus(\mathbf{e}_{t}))
\end{equation}
where $LN(\cdot)$ is layer normalization.

\subsubsection{Down Projector}
In the down projector, we first conduct element-wise product to fuse $\mathbf{d}_{t}$ and text feature vector $\mathbf{h}_{1}$.
Then, a stack of linear layers followed by softmax and layer normalization are leveraged to predict the noise $\epsilon_{\theta}$.
\begin{equation}
    \epsilon_{\theta}= projector(\mathbf{d}_{t}, \mathbf{h}_{1})
\end{equation}

\begin{table}[H]
    \centering
    \caption{Detailed statistics of the datasets.}\label{stat}
    \resizebox{\linewidth}{!}{
        \begin{tabular}{cccc}
            \toprule[1pt]
            \textbf{Dataset} & \textbf{Training Set} & \textbf{Test Set} & \textbf{\#Avg. words}  \\ \midrule
            \textbf{AG NEWS}   & 120K   & 7.6K   & 43          \\
            \textbf{SST-2} & 67K  & 1.8K  & 19           \\
            \textbf{MRPC} & 3.7K & 1.7K & 44 \\
            \bottomrule[1pt]
        \end{tabular}
    }
\end{table}

\section{Experiments}

\begin{table*}[ht]
\renewcommand\arraystretch{1}
\setlength\tabcolsep{6pt}
\centering
\caption{Experimental results of adversarial robustness evaluation. The best performance is marked in \textbf{bold}. For the MRPC task, we attack both hypothesis and premise.  
Methods labelled by  $\dag$ are fine-tuning baselines without considering adversarial defence. 
}
\label{tab:main results}
\scalebox{0.9}{
\begin{tabular}{c|l|c|c|c|cc}
\toprule[1pt]
\multicolumn{1}{c|}{\multirow{2}{*}{\textbf{Dataset}}} &
\multicolumn{1}{c|}{\multirow{2}{*}{\textbf{Method}}} &
\multicolumn{1}{c|}{\multirow{2}{*}{\textbf{Clean$\%$}}} &
\multicolumn{2}{c|}{\textbf{TextFooler}} & 
\multicolumn{2}{c}{\textbf{BERT-Attack}} \\ \cline{4-7}
\multicolumn{1}{c|}{} & \multicolumn{1}{c|}{} & \multicolumn{1}{c|}{}& 
\multicolumn{1}{c}{\textbf{Aua$\%$}} &
\multicolumn{1}{c|}{\textbf{Suc$\%$} } & 
\multicolumn{1}{c}{\textbf{Aua$\%$} } & 
\multicolumn{1}{c}{\textbf{Suc$\%$} } \\ \cline{1-7}
\hline
\multirow{7}{*}{\textbf{AG NEWS}}
& BERT$^{\dag}$ & $94.0$ & $20.5$  & $78.9$  & $14.6$  & $84.3$ \\ 
& PGD~\cite{litoken} & $94.8$ & $36.2$  & $61.3$ & $32.8$  & $65.7$ \\
&Free LB~\cite{Zhu2020FreeLB} &$94.7$ & $34.8$ & $63.5$  & $12.7$  & $86.7$ \\
&InfoBERT~\cite{pmlr-v119-ishida20a} & $94.9$ & $30.4$  & $65.9$  & $20.4$  & $78.0$  \\
&Text Purification~\cite{li-etal-2023-text} & $93.0$ & $51.0$  & 42.0 & $44.5$ &48.5 \\
&\textbf{ROIC-DM} & $\mathbf{ 95.1}$ & $\mathbf{78.7 }$  & $\mathbf{18.4 }$   & $\textbf{49.0 }$  & $\textbf{ 47.0}$ \\
\hline
\multirow{7}{*}{\textbf{SST-2}}
&BERT$^{\dag}$& $92.2$ & $13.9$  & $80.4$ & $13.3$  & $80.6$ \\ 
&PGD& $93.2$ & $13.4$  & $82.1$ & $13.4$  & $84.5$ \\
&Free LB & $92.2$ & $19.4$  & $80.3$ & $12.1$  & $87.1$ \\
&InfoBERT & $92.9$ & $20.4$  & $76.7$ & $16.6$ & $82.7$ \\
&Text Purification & $91.8$ & $42.6$ & $53.4$ & $33.5$  & $62.4$\\
&\textbf{ROIC-DM} & $\mathbf{94.1} $ & $\mathbf{ 55.4}$  & $\mathbf{39.5} $   & $\mathbf{ 46.4}$  & $ \mathbf{49.7}$ \\
\hline
\multirow{7}{*}{\textbf{MRPC}}
&BERT$^{\dag}$ & $83.8$ & $6.4$  & $92.8$ & $9.4$  & $89.5$ \\ 
&PGD & $84.3$ & $6.9$  & $92.2$ & $11.5$  & $82.3$ \\
&Free LB& $83.8$ & $8.2$  & $ 91.0$  & $10.3$ & $87.7$ \\
&InfoBERT & $87.9$ & $13.9$  & $84.1$ & $17.4$  & $78.9$ \\
&Text Purification & $82.5$ &  $32.5$  & $54.5$  & $28.6$  & $61.2$\\
&\textbf{ROIC-DM} & $\mathbf{90.8} $ & $ \mathbf{53.3}$  & $\mathbf{39.8 }$   & $ \mathbf{38.3}$  & $\mathbf{58.5} $ \\
\bottomrule[1pt]
\end{tabular}
}
\end{table*}
\begin{table*}[!htbp]
\renewcommand\arraystretch{1.2}
\setlength\tabcolsep{6pt}
\centering
\caption{The comparison of the accuracy of ROIC-DM with different pre-trained advisors on the AG NEWS dataset.}
\label{ablation3}
\scalebox{0.9}{
\begin{tabular}{l|l|l|l|l|l}
\toprule[1pt]
\multirow{2}{*}{\textbf{Method}} & \multirow{2}{*}{\textbf{Accuracy \%}}&\multirow{2}{*}{\textbf{Method}} & \multirow{2}{*}{\textbf{Accuracy \%}}&\multirow{2}{*}{\textbf{Method}} & \multirow{2}{*}{\textbf{Accuracy \%}} \\
& &&& \\
\midrule
BERT & 94.0&DistilBERT&93.4&ALBERT &94.5 \\
ROIC-DM(-advisor) & 91.8 &ROIC-DM(-advisor)&91.8 &ROIC-DM(-advisor)&91.8\\
\textbf{ROIC-DM} & \textbf{95.1}&\textbf{ROIC-DM}&\textbf{93.8}&\textbf{ROIC-DM}&\textbf{95.5} \\
\bottomrule[1pt]
\end{tabular}
}
\end{table*}

In this section, we first introduce the basic experimental settings, and then, present the experimental results with comprehensive analysis to showcase the superiority of our proposed methods.


\subsection{Datasets}
In this paper, we conduct experiments on three widely used text classification and inference datasets: AG NEWS~\cite{NIPS2015_250cf8b5}, SST-2~\cite{socher-etal-2013-recursive}, and MRPC~\cite{dolan-brockett-2005-automatically}.
The statistics of these involved datasets are illustrated in Table~\ref{stat}, including the size of the training/test set and the average word count of the training samples. 
The training and test set division is following~\cite{liu2022flooding}.
We use the whole test set to evaluate model accuracy and randomly select $500$ samples for robustness evaluation since the attack process is seriously slow~\cite{morris-etal-2020-textattack,moon2023randomized}.

\subsection{Baselines}\label{baseline}
We compare our ROIC-DM with the following defense baselines including two adversarial training algorithms (PGD and FreeLB), one regularization method (InfoBERT), and a text purification method. 
\subsubsection{PGD}~\cite{madry2018towards} formulates adversarial training as a minimax problem, which aims to minimize the empirical loss on adversarial examples that could potentially lead to adversarial risk.
\subsubsection{FreeLB}~\cite{Zhu2020FreeLB} attempts to improve language models' robustness by enhancing their generalization abilities. Specifically, FreeLB generates some virtual adversarial samples by injecting adversarial perturbations into word embeddings. Then, FreeLB mixes them with normal training data to improve the tolerance of target models to adversarial samples.
\subsubsection{InfoBERT}~\cite{wang2021infobert} consists of two mutual-information-based regularizers to improve the robustness of the learned representations by suppressing noisy mutual information.
\subsubsection{Text Purification}~\cite{li-etal-2023-text} is a textual adversarial purification algorithm. It utilizes the mask-infill ability of pre-trained models to recover noisy texts and use these purified texts to make predictions.

\subsection{Attack Methods and Evaluation Metrics}\label{metric}
Textfooler~\cite{jin2020bert} and BERT-Attack~\cite{li2020bert} have exhibited capable of effectively deceiving robust models in text classification and inference tasks with limited perturbations.
These two attack methods have been widely used in adversarial robustness research~\cite{liu2022flooding,morris-etal-2020-textattack}.
Therefore, in this paper, we leverage these two adversarial attack methods to assess the robustness of our ROIC-DM. 

TextFooler identifies crucial words within the input text for the target model and iteratively replaces them with synonyms until the model's prediction is modified.
BERTAttack utilizes BERT in a manner that preserves semantics when generating substitute words for the identified vulnerable words in the input text.

We utilize the following metrics to evaluate models' resistance to above mentioned adversarial attacks:

\subsubsection{Clean\%} is the victim model's accuracy tested on the clean test set, which represents the original performance of the victim model.
\subsubsection{Aua\%} is short for  \emph{a}ccuracy \emph{u}nder \emph{a}ttacks(Aua). 
It measures the prediction accuracy of victim models on the adversarial data generated by certain attacks.
This metric reflects the defensive capability of the model against adversarial attacks. A higher $Aua\%$ indicates a more robust model.
\subsubsection{Suc\%} is the attack success rate. i.e., it is the ratio of the number of texts that have been successfully perturbed by certain attack methods to the number of involved texts.
A lower $Suc\%$ means a more robust model.

\subsection{Implementation Details}\label{implement}
For ROIC-DM, we set the number of diffusion timesteps to $T = 1000$, and employ a linear noise schedule with $\beta_1 =1e-4$ and $\beta_{T} = 0.02$. The optimizer is AdamW~\cite{loshchilov2017decoupled} with a linear decay learning rate starting at 1e-4.
The batch size is 64. 
For the pre-trained advisor, we directly use the pre-trained version in the toolkit~\cite{morris-etal-2020-textattack} provided by Huggingface\footnote{\url{https://huggingface.co/textattack}}.

For the baselines, we directly use the hyper-parameter settings in their released codes (PGD\footnote{\url{https://github.com/MadryLab/mnist_challenge}}, FreeLB\footnote{\url{https://github.com/zhuchen03/FreeLB}}, InforBERT\footnote{\url{https://github.com/AI-secure/InfoBERT}}) to generate experimental results. For Text purification~\cite{li-etal-2023-text}, we re-implement it according to their paper description since there is no public code available.


For the adversarial attack methods, we directly use the corresponding attack's official codes in Openattack~\cite{zeng2020openattack} framework.

All the code related to the experiments will be available at \url{https://github.com/} after the double-blind review.

\subsection{The Robustness of ROIC-DM}\label{ex-result}
In this part, we showcase that our ROIC-DM outperforms traditional language models in robustness, even when the latter are equipped with advanced defense methods.

Table~\ref{tab:main results} presents the experimental results of ROIC-DM compared with BERT equipped with baseline defense methods. Note that our ROIC-DM also utilizes BERT as the advisor for the fair comparison.
According to the results on clean data ($Clean\%$), we can observe that ROIC-DM outperforms its advisor, BERT, on all three datasets for at most $2.9$ accuracy scores ($Clean\%$). Besides, ROIC-DM is also better than BERT with defenses on the clean data.

\begin{table}[!htbp]
\renewcommand\arraystretch{1.2}
\setlength\tabcolsep{6pt}
\centering
\caption{Comparison of the performance between ROIC-DM(-advisor) and ROIC-DM on the AG News test dataset under BERTAttack and Textfooler adversarial attacks. BERT is used as the advisor.}
\label{tab:ablation}
\resizebox{\linewidth}{!}{
\begin{tabular}{c|l|cc|cc}
\toprule[1pt]
\multirow{2}{*}{\textbf{Method}}  & \multirow{2}{*}{\textbf{Clean \%}} & \multicolumn{2}{c|}{\textbf{TextFooler}} & \multicolumn{2}{c}{\textbf{BERT-Attack}} \\
\cline{3-6}
 & & \textbf{Aua \%} & \textbf{Suc \%} & \textbf{Aua \%} & \textbf{Suc \%} \\
\midrule
ROIC-DM(-advisor) & 91.8 & 60.3& 33.4& 46.4&47.8 \\
\textbf{ROIC-DM} & \textbf{95.1} & \textbf{78.7} & \textbf{18.4} & \textbf{49.0} & \textbf{47.0} \\
\bottomrule[1pt]
\end{tabular}
}
\end{table}

\begin{table*}[!htbp]
\renewcommand\arraystretch{1.2}
\setlength\tabcolsep{6pt}
\centering
\caption{A case study on the AG NEWS dataset}\label{tb_case}
\begin{tabular}{l|p{7cm}|l|l|l}
\toprule[1pt]
\multicolumn{1}{c|}{\textbf{}} & \multicolumn{1}{c|}{\textbf{Text}} & \multicolumn{1}{c|}{\textbf{BERT}} & \multicolumn{1}{c|}{\textbf{Text Purification}} & \multicolumn{1}{c}{\textbf{Ours}} \\ \midrule
\multirow{3}{*}{\textbf{Original}}                      &         though the violence is far less sadistic than usual, the film is typical miike: fast, furious and full of off-the-cuff imaginative flourishes                       &                  \multirow{3}{*}  { $\checkmark   $ }       &     \multirow{3}{*}  {$ \checkmark  $     }                 &     \multirow{3}{*}{\checkmark  }                        \\ \hline
\multirow{3}{*}{\textbf{Attacked by BERT-Attack} }                     &      since the brutal is very smaller sad graphic than usual, the movie is traditional miike way rapid becomes frantic and - of off the cuff creative flourishs.                       &        \multirow{3}{*}  {$  \times   $  }               &                      \multirow{3}{*} {$\times   $ }       &             \multirow{3}{*} {$\checkmark   $ }                  \\ \bottomrule[1pt]
\end{tabular}
\end{table*}

When attacked by TextFooler, the vanilla BERT's performance is dramatically dropped on all three datasets. Text Purification achieves the best performance among baselines, but our ROIC-DM still outperforms it by a large margin (e.g., 27.7 $Aua\%$ scores on AG NEWS). Moreover, it is worth mentioning that, Text Purification is harmful to the model's performance on clean data as shown in $Clean\%$.

BERT-Attack is stronger than TextFooler and all the models suffer performance deterioration but our ROIC-DM still consistently outperforms all these baseline methods.

\subsection{ROIC-DM v.s. Traditional Classifiers}
Except for the robustness, another advantage of ROIC-DM is that it can incorporate knowledge from traditional classifiers and achieve better performance. 

Table~\ref{ablation3} presents the results of ROIC-DM using different pre-trained advisors. Due to the space limitation, we only include the results on AG NEWS. A similar conclusion can be obtained from the other two datasets. As shown in the results, when removing the advisor, ROIC-DM's performance is slightly worse than traditional classifiers. This may be because these classifiers are based on pre-trained models that have been trained on large-scale datasets with long-term developed model architecture, while our ROIC-DM is a new kind of classification model trained from scratch.

When ROIC-DM uses counterpart advisors, it achieves better performance than its advisors. Specifically, ROIC-DM obtains $1.1$, $0.4$, and $1.0$ higher scores compared to BERT, DistillBERT, and ALBERT, respectively.

\begin{figure}[t]
\centering
\includegraphics[width=0.9\columnwidth]{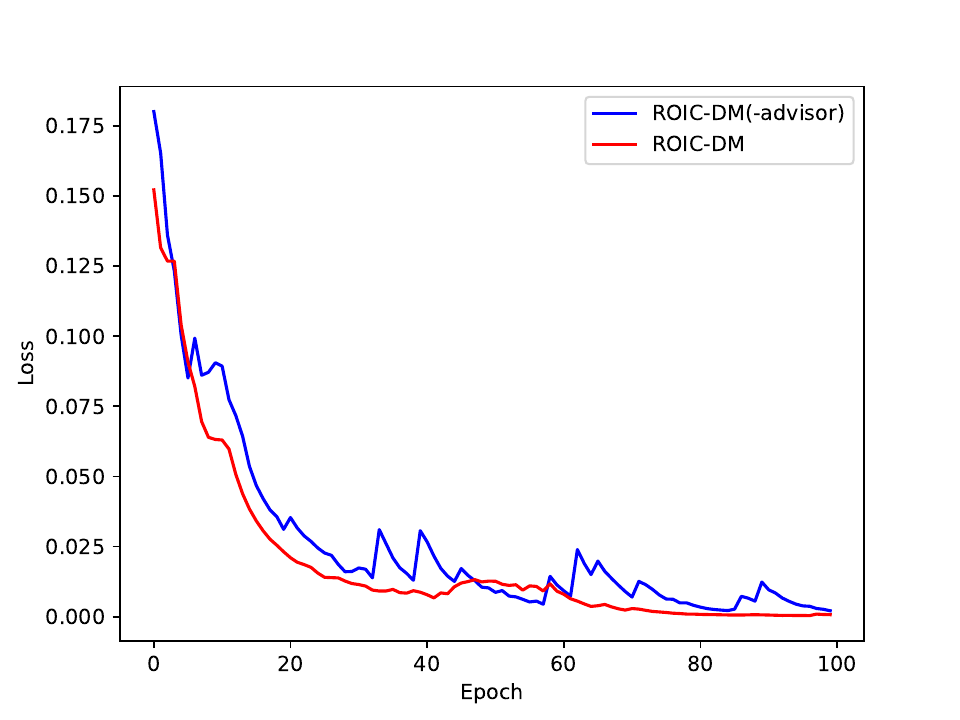} 
\caption{The training loss trend for ROIC-DM and ROIC-DM(-advisor) on the AG NEWS dataset.}
\label{loss curve}
\end{figure}

\subsection{The Impact of Advisor}\label{ablation}
In this section, we investigate the impacts of advisors for ROIC-DM from the robustness perspective and the training aspect.
We compare ROIC-DM with BERT as advisor and ROIC-DM(-advisor) on the AG News dataset under BertAttack and Textfooler adversarial
attacks.
As displayed in Table~\ref{tab:ablation}, without an advisor, ROIC-DM's performance declined on both clean and adversarial data. However, if combined the results in Table~\ref{tab:ablation} and Table~\ref{tab:main results}, we can observe that without an advisor, ROIC-DM still exhibits strong robustness, as its performance ($60.3$ and $46.4$ with $Aua\%$ for TextFooler and BERT-Attack respectively) is still much better than BERT with the best baseline defender ($51.0$ and $44.5$ for TextFooler and BERT-Attack respectively). This observation supports our argument that the diffusion model will be more robust than conventional language models.

From the training aspects, Figure~\ref{loss curve} shows the training loss curve for ROIC-DM and ROIC-DM(-advisor). The training loss curve for ROIC-DM is smoother than ROIC-DM(-advisor) and the prior's decreasing speed is also faster than the latter. This phenomenon implies that the advisor can be positive to ROIC-DM's training process.

\subsection{Case Study}
Table~\ref{tb_case} presents the data sample from the AG NEWS dataset. We utilize BERT-Attack to perturb the original data since it is one of the strongest textual adversarial attacks. We show the comparison with Text Purification as it has the best performance among baselines. For the original text, the vanilla BERT, Text Purification based BERT, and our ROIC-DM can make the correct prediction. After being perturbed, the overall meaning of the sentence is the same as the original text but BERT and Text Purification based BERT cannot make correct predictions. Only our ROIC-DM keeps correct, which indicates its robustness.

\section{Conclusion}
In this paper, we introduce an innovative model for robust text inference and classification named ROIC-DM, which is built upon the foundational framework of the diffusion model. To enhance the performance of ROIC-DM, we strategically integrate conventional large language models as advisors within the reverse process. The experimental results on three datasets with two competitive textual adversarial attacks indicate that ROIC-DM is more robust to adversarial attacks and can achieve better performance compared with conventional language models. 

\bigskip

\bibliography{aaai24}

\end{document}